%% file: arxiv.tex
\pdfoutput=1
\documentclass{article}
\usepackage{pam_preprint,times}

\input{preamble}
\arxivtrue

\author{Fangyuan Wang$^{1}$, Songhao Huang$^{1}$, Haoxiang Sun$^{1}$, Shipeng Lyu$^{1}$, Chengyang He$^{2}$, \\
\textbf{Anqing Duan$^{3}$, Peng Zhou$^{4}$, David Navarro-Alarcon$^{1,}$\thanks{Corresponding author.}} \\
$^{1}$Hong Kong Polytechnic University \quad
$^{2}$National University of Singapore \\
$^{3}$Mohamed bin Zayed University of Artificial Intelligence \quad
$^{4}$Great Bay University}

\hypersetup{
  pdftitle={Proprioceptive Sketches as Long-Horizon Intent for Generative Action Policies},
  pdfauthor={Fangyuan Wang, Songhao Huang, Haoxiang Sun, Shipeng Lyu, Chengyang He, Anqing Duan, Peng Zhou, David Navarro-Alarcon},
}

\begin{document}

\maketitle

\input{content/teaser}

\input{content/abstract}

\input{content/body}

\subsection*{Reproducibility Statement}
Appendix~\ref{app:spec} describes sketch construction, temporal invariance, and reconstruction audits. Appendix~\ref{app:arch} specifies the model, training objectives, and sampling procedures. Appendix~\ref{app:configs} collects training configurations, dataset splits, evaluation protocols, and the sketch intervention procedure.

\bibliography{main}
\bibliographystyle{pam_preprint}

\appendix
\input{content/appendix}

\end{document}

%% file: preamble.tex
\usepackage{hyperref}
\usepackage{url}
\usepackage{amsmath,amssymb}
\usepackage{amsthm}
\usepackage{algorithm}
\usepackage{algpseudocode}
\algrenewcommand\algorithmicrequire{\textbf{Input:}}
\algrenewcommand\algorithmicensure{\textbf{Output:}}
\usepackage{booktabs}
\usepackage{subcaption}
\usepackage{xcolor}
\usepackage{colortbl}
\usepackage{tabularx}
\usepackage{graphicx}
\usepackage{wrapfig}
\usepackage{needspace}
\usepackage{tikz}

\newif\ifarxiv

\title{Proprioceptive Sketches as Long-Horizon Intent for Generative Action Policies}

%% file: content/teaser.tex
\begin{figure}[h]
  \centering
  \includegraphics[width=\textwidth]{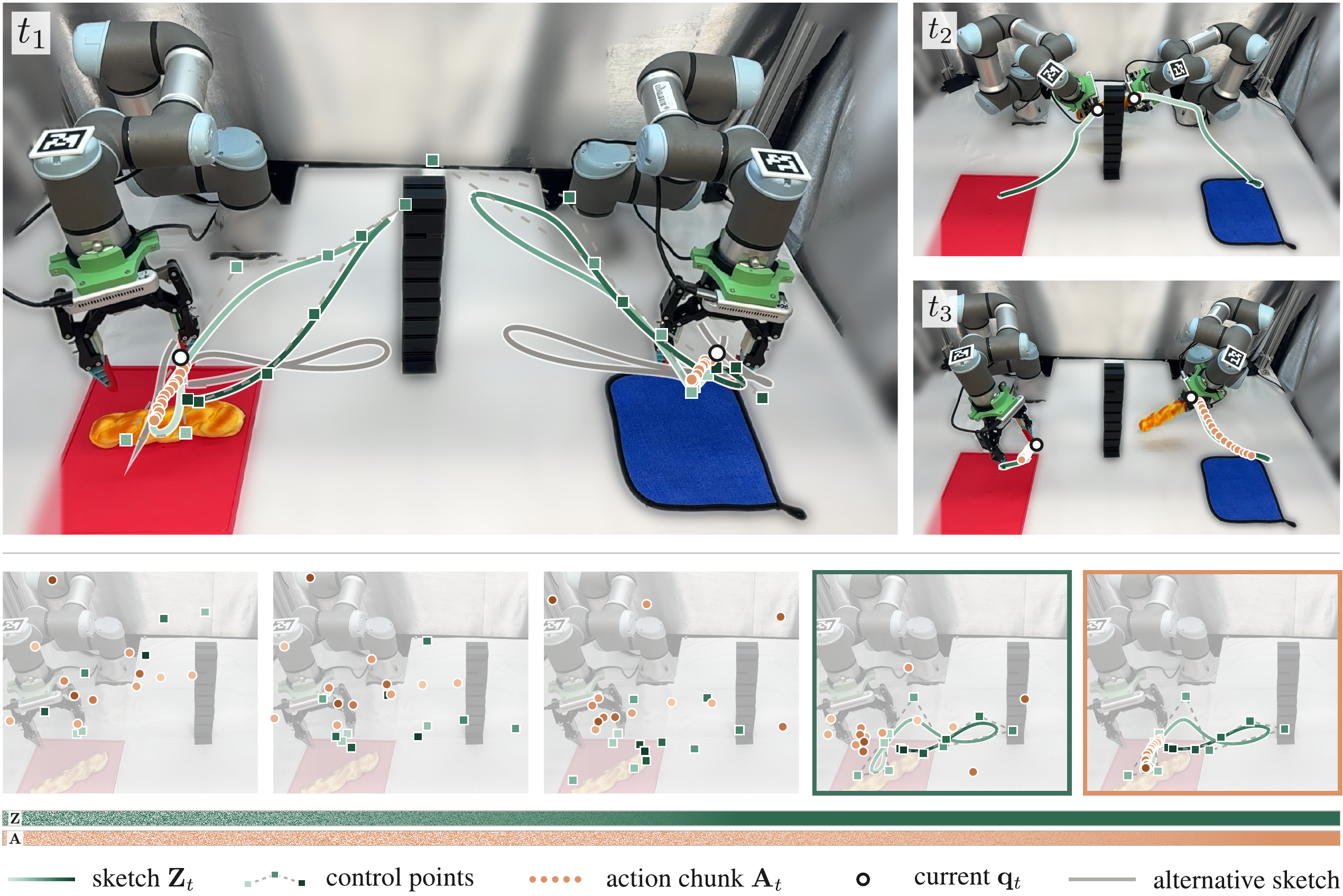}
  \caption{\textbf{PAM replans during a real-world dual-arm long-horizon task.} At each step, PAM jointly generates a proprioceptive sketch $\mathbf Z_t$ of the future configuration-space
path and a short action chunk $\mathbf A_t$ conditioned on the sketch, while only $\mathbf A_t$ is executed. For visualization, the sketch and the action chunk are mapped through forward
kinematics and projected into the camera view.}
  \label{fig:teaser}
\end{figure}

%% file: content/abstract.tex
\begin{abstract}
  Generative robot policies predict short action chunks but lack explicit long-horizon intent. Recent methods expose longer-horizon structure through language plans, subgoal images, or video forecasts, which are costly to generate and still need to be translated into robot motion. Predicting future robot motions avoids this translation, but a dense, time-indexed trajectory requires numerous parameters to cover the full remaining task, and over a short horizon it largely repeats the action chunk and adds little guidance for action generation. We propose Proprioceptive Action Models (PAM), which jointly generate a compact, timing-free sketch of the robot's remaining joint-space path and a dense executable action chunk within a single transformer denoiser. The sketch parameterizes the path by arc length rather than time, capturing geometric intent invariant to execution timing. Block-causal attention and a staggered denoising schedule maintain directed sketch-to-action dependence, ensuring the action tokens condition on a progressively cleaner sketch throughout sampling. In simulation, PAM improves over its action-only counterparts on Push-T and LIBERO-Long; on four real-world bimanual tasks, it raises success from 47.5\% to 75.0\%.\ifarxiv{} Project page: \url{https://nicehiro.github.io/pam_dp/}.\fi
\end{abstract}

%% file: content/body.tex
\section{Introduction}
\label{sec:intro}
Generative action policies have become a dominant paradigm for visuomotor robot control, producing temporally coherent commands through action chunking~\citep{zhaoACTBimanual2023,chiDiffusionPolicy2023,blackVisionLanguageActionFlow2024,reussFLOWERDemocratizingGeneralist2025}. Despite strong performance on short-horizon tasks, these policies remain less effective on long-horizon manipulation~\citep{fanLongVLA2025}. One approach to this challenge is to guide action prediction with visual forecasts~\citep{black2023zero,huVideoPredictionPolicy2025,tianPredictiveInverseDynamics2024}, predictive scene latents~\citep{zhangDreamVLAVisionLanguageAction2025,liuLaST02026,zhuUnifiedWorldModels2025}, or language-based plans and reasoning~\citep{wenDiffusionVLAScalingRobot2024,zawalskiRoboticControlEmbodied2025}. These representations capture anticipated scene evolution or semantic task structure, but can be costly to generate or require additional learned mappings to continuous control. This motivates a compact representation of long-horizon intent closely tied to the robot’s own motion.

Future robot configurations provide a natural representation of the robot's intended motion. Demonstrations contain synchronized proprioceptive measurements, providing supervision without additional annotation in a space directly related to robot control. Prior work has used robot-motion representations to guide action prediction, including image-plane trajectories~\citep{guRTTrajectory2023,niuLLARVA2024} and predicted end-effector paths~\citep{chenOASIS2026}. However, a dense, time-indexed configuration trajectory over the full remaining task is long and varies in length across demonstrations, making it difficult to represent with a limited number of parameters. Restricting the prediction to a short horizon instead makes it largely redundant with the executable action chunk, so it provides little additional guidance for action generation. An effective intent representation should therefore cover the full remaining motion with a fixed number of parameters while providing information beyond the action chunk.

We introduce the \emph{proprioceptive sketch}, a compact representation of the robot’s remaining configuration-space path to task completion. The sketch uses arc-length parameterization to remove execution timing and a small number of B-spline coefficients to limit geometric details. In the handoff task in Figure~\ref{fig:teaser}, the sketch describes the arm paths toward the transfer region and around the barrier without specifying motion speed or the duration of the exchange. It provides long-horizon geometric context for predicting a separate, short-horizon action chunk. Only the action chunk is executed, determining the immediate motion and gripper commands.

\textbf{Proprioceptive Action Models (PAM)} jointly generate the sketch and an action chunk within a single transformer denoiser. As illustrated in Figure~\ref{fig:teaser}, both outputs are initialized with Gaussian noise and refined through iterative denoising. A block-causal attention mask allows action tokens to attend to the sketch while preventing sketch tokens from reading the action stream. Following Diffusion Forcing~\citep{chenDiffusionForcing2024}, we train the denoiser with varying corruption levels across the two streams and use a staggered denoising schedule that denoises the sketch ahead of the actions, providing cleaner geometric context for action generation.

Our contributions are threefold. First, we formulate the proprioceptive sketch, a compact, timing-free representation of the remaining configuration-space path that serves as an explicit intermediate variable for action generation and is, by construction, invariant to temporal reparameterization of the demonstrated path. Second, we develop a joint generator with directed attention and staggered denoising to guide action prediction with the sketch. Third, we apply PAM to a diffusion transformer trained from scratch (PAM-DP), and a pretrained vision-language-action model (PAM-VLA). We evaluate PAM-DP on Push-T and four real-world bimanual tasks, and PAM-VLA on LIBERO-Long.

\section{Related Work}
\label{sec:related_works}

\textbf{Predictive intermediate representations.} Intermediate predictions provide action policies with information about intended behavior. Language-based approaches express task structure through subtask descriptions or reasoning traces~\citep{wenDiffusionVLAScalingRobot2024,black2025pi05,zawalskiRoboticControlEmbodied2025}. Visual approaches use future images or video to represent goal configurations and scene evolution~\citep{black2023zero,zhaoCoTVLAVisualChainofThought2025,huVideoPredictionPolicy2025,tianPredictiveInverseDynamics2024}. Other methods learn predictive representations of scene dynamics or visuomotor behavior in latent space~\citep{zhangDreamVLAVisionLanguageAction2025,liuLaST02026,liang2024skilldiffuser,xuFutureVLA2026}. These representations describe the task rather than the robot's own motion, and they are either expensive to generate or require a learned decoder to reach continuous control. PAM instead predicts intended motion as a configuration-space path, which requires no annotation beyond recorded proprioception.

\textbf{Robot-motion representations.} Robot motion provides an intermediate representation for planning and control. Hierarchical reinforcement learning uses state-space subgoals to guide lower-level policies~\citep{nachum2018data}. RT-Trajectory conditions action prediction on image-plane trajectory sketches~\citep{guRTTrajectory2023}, while TraceVLA supplies visual traces of past motion~\citep{zhengTraceVLAVisualTrace2024}. Predictive approaches use future motion in different ways. GeoPredict supervises future keypoint prediction during training~\citep{qianGeoPredictLeveragingPredictive2025}, whereas OASIS predicts end-effector trajectories that condition a separate action decoder~\citep{chenOASIS2026}. Recent spline policies similarly parameterize executable action trajectories with spline coefficients~\citep{tianSplinePolicy2025,hanBSplinePolicy2026,yangABPolicy2026}. PAM likewise predicts future robot motion but differs in three respects: the sketch spans the remaining path to task completion, it removes execution timing through arc-length parameterization, and it is generated jointly with actions in a single denoiser.

\textbf{Joint generation and denoising schedules.} Joint world-action models generate actions together with predicted images, video, or scene representations~\citep{zhuUnifiedWorldModels2025,yeWorldActionModels2026,linHiFVLA2026,xuActionGeometry2026}. Trajectory-diffusion methods instead generate state-action plans~\citep{janner2022planning} from which actions are recovered through inverse dynamics~\citep{ajay2023conditional}. Diffusion Forcing introduces independent token-level corruption during training, enabling flexible denoising schedules at inference~\citep{chenDiffusionForcing2024}. PAM draws on these joint-generation and scheduling principles to couple a non-executable motion sketch with an executable action chunk. Directed attention allows actions to depend on the sketch without feeding action information back into sketch generation, while staggered denoising lets the sketch become cleaner ahead of the actions.

\section{Method}
\label{sec:method}

\subsection{Problem Setup}
We consider demonstration trajectories $\tau=\bigl(\ell,\{(o_t,\mathbf q_t,\mathbf a_t)\}_{t=1}^{T}\bigr)$, where $\ell$ is an optional task specification and $T$ is the trajectory length. At time $t$, $o_t$ denotes the policy observation, $\mathbf q_t\in\mathbb R^{d_q}$ is the robot configuration, and $\mathbf a_t\in\mathbb R^{d_a}$ is the action. The short-horizon action target is $\mathbf A_t=[\mathbf a_t,\ldots,\mathbf a_{t+H-1}]\in\mathbb R^{H\times d_a}$, where $H$ is the prediction horizon. We additionally introduce a proprioceptive sketch $\mathbf Z_t$ of the remaining configuration path $\mathbf q_{t:T}$.
PAM models the joint policy
\begin{equation}
\label{eq:policy}
(\mathbf Z_t,\mathbf A_t)\sim\pi_\theta(\,\cdot\mid o_t,\mathbf q_t,\ell).
\end{equation}
\textbf{Denoising generative policies.} We instantiate $\pi_\theta$ as a conditional denoising generative model. For a clean normalized token $\mathbf x^0$ and a noise coordinate $\gamma\in[0,1]$, with $\gamma=0$ denoting clean data and $\gamma=1$ pure noise, the corrupted token is
\begin{equation}
\label{eq:corruption}
\mathbf x^\gamma=\alpha(\gamma)\mathbf x^0+\sigma(\gamma)\boldsymbol\epsilon,\qquad\boldsymbol\epsilon\sim\mathcal N(\mathbf 0,\mathbf I),
\end{equation}
where the schedule $(\alpha,\sigma)$ is set by the generative formulation. A denoiser $f_\theta$ takes corrupted tokens, their noise coordinates, and the conditioning $\mathbf c_t=(o_t,\mathbf q_t,\ell)$, and predicts a regression target $\mathbf u$: the noise $\boldsymbol\epsilon$ in diffusion models~\citep{hoDDPM2020} or the velocity $\boldsymbol\epsilon-\mathbf x^0$ in rectified flow~\citep{liuRectifiedFlow2023}. Sampling starts from Gaussian noise and applies $K$ reverse updates, such as DDIM~\citep{songDDIM2021} or Euler steps, that move $\gamma$ from one to zero. PAM applies this procedure jointly to two token streams, the sketch $\mathbf Z_t$ (Section~\ref{sec:sketch}) and the action chunk $\mathbf A_t$, with a single transformer denoiser (Section~\ref{sec:joint}).

\subsection{The Proprioceptive Sketch}
\label{sec:sketch}
\begin{wrapfigure}[15]{r}{0.5\textwidth}
  \centering
  \includegraphics[width=0.5\textwidth]{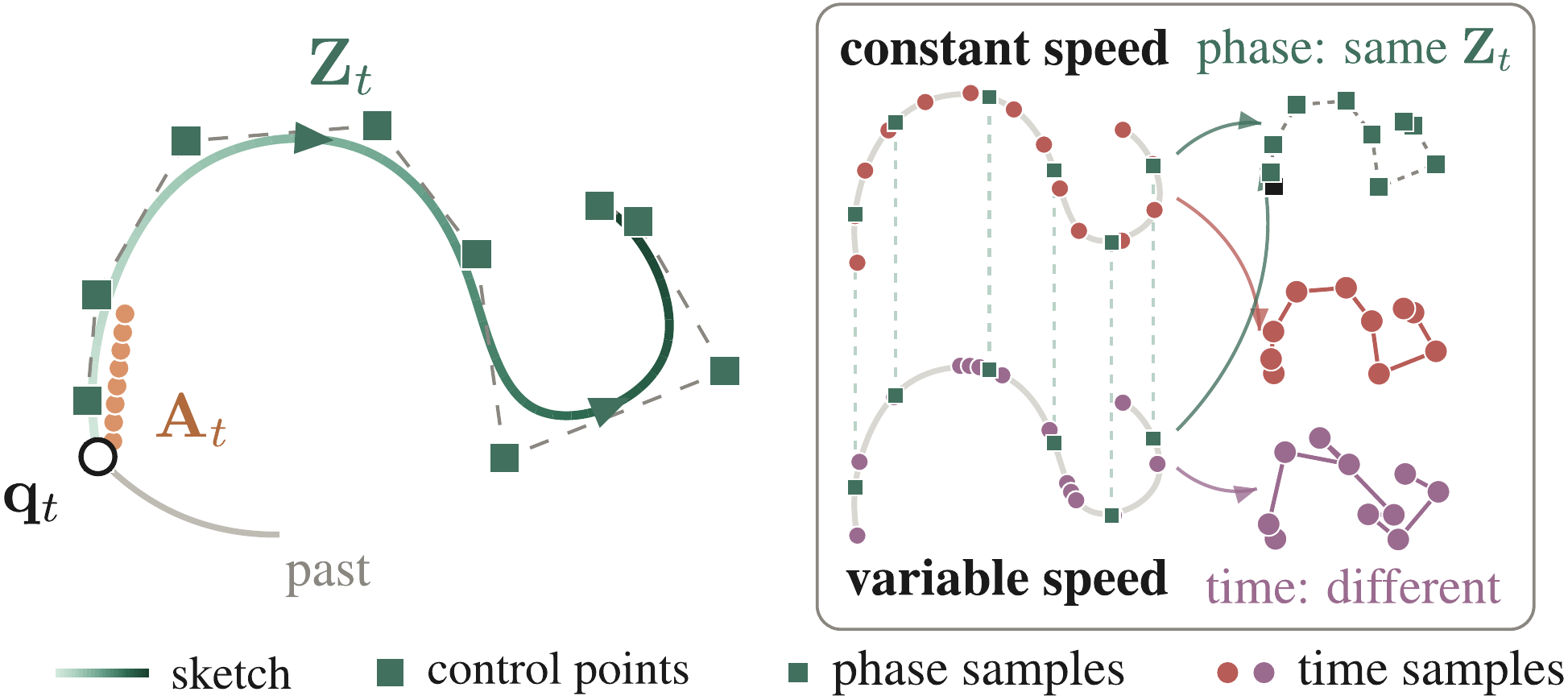}
  \caption{The proprioceptive sketch. Left, B-spline coefficients $\mathbf Z_t$ represent the remaining configuration path, while only the action chunk $\mathbf A_t$ is executed. Right, re-timing the same continuous path changes its time-indexed representation but preserves its phase-indexed sketch.}
  \label{fig:sketch}
\end{wrapfigure}

The proprioceptive sketch represents the geometry of the remaining configuration path without specifying its execution timing (Figure~\ref{fig:sketch}). We parameterize the path by arc length and encode it with several B-spline coefficients.

\textbf{Phase.}
For a remaining path $\mathbf q_{t:T}$, we define cumulative arc length and normalized phase as
\begin{equation}
\label{eq:phase}
s_u=\sum_{v=t}^{u-1}
\lVert\mathbf q_{v+1}-\mathbf q_v\rVert_2,
\qquad
\phi_u=\frac{s_u}{L},
\end{equation}

for $u\in\{t,\ldots,T\}$ and $L > 0$. Here, $s_u$ is the cumulative configuration-space distance from time $t$ to $u$, and $L=s_T$ is the total length of the remaining path. Phase measures the fraction of path length traversed rather than elapsed time. Stationary intervals therefore produce no phase increment. When $L=0$, we define the sketch to be zero.

\textbf{Representation.}
We encode the phase-indexed path with a clamped cubic B-spline using $M\geq4$ control points and uniformly spaced interior knots. The sketch is the coefficient matrix $\mathbf Z_t=[\mathbf z_{t,0},\ldots,\mathbf z_{t,M-1}]^\top\in\mathbb R^{M\times d_q}$, which decodes to a relative configuration path
\begin{equation}
\label{eq:sketch}
\Delta\hat{\mathbf q}_t(\phi)
=
\sum_{m=0}^{M-1}B_{m,3}(\phi)\,\mathbf z_{t,m},
\phi\in[0,1].
\end{equation}
where $B_{m,3}$ denotes a cubic B-spline basis function. The corresponding absolute path is $\hat{\mathbf q}_t(\phi)=\mathbf q_t+\Delta\hat{\mathbf q}_t(\phi)$. We fix $\mathbf z_{t,0}=\mathbf 0$ so that the path begins at the current configuration. PAM generates only the remaining $M-1$ coefficients regardless of the length of the remaining path.

\textbf{Training target and representation capacity.} We construct sketch targets by fitting the remaining configuration path in each demonstration. We first express the path relative to $\mathbf q_t$ and resample it uniformly in arc-length phase. We then fit the B-spline by least squares, fixing the first and last control points to $\mathbf 0$ and $\mathbf q_T-\mathbf q_t$, respectively. All control points except the fixed zero anchor form the $M-1$ prediction target. The number of control points $M$ controls the trade-off between compactness and geometric fidelity. We use the RMS per-step configuration change in the training demonstrations as a reference for reconstruction accuracy. The adopted coefficient budgets keep the spline reconstruction RMSE below this reference for each task. Appendix~\ref{app:spec} details the knot construction, fitting procedure, normalization, and capacity audit.

\textbf{Temporal invariance.}
Changing the speed of traversal along a fixed continuous path does not change its arc-length parameterization. The sketch therefore remains unchanged under temporal reparameterization, including pauses, for any fixed coefficient budget $M$. This property concerns the underlying continuous path; finite sampling and measurement noise can introduce differences between separately recorded demonstrations.

\subsection{Joint Sketch and Action Generation}
\label{sec:joint}
\begin{figure}[t]
  \centering
  \includegraphics[width=\textwidth]{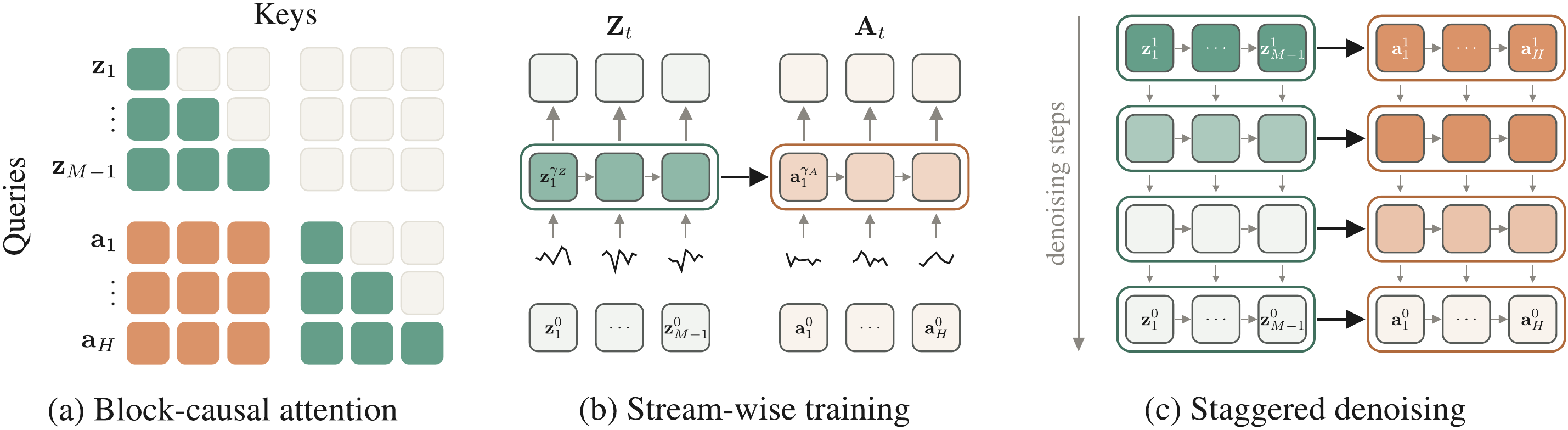}
  \caption{Joint sketch and action generation. (a) Block-causal attention allows actions to read the sketch while preventing information flow in the reverse direction. (b) Training assigns independent noise levels to the sketch and action streams, with a shared level within each stream. (c) Staggered denoising advances the sketch ahead of the action chunk.}
  \label{fig:method}
\end{figure}

PAM jointly generates the sketch and action chunk with a single transformer denoiser $f_\theta$~\citep{peebles2023scalable}. The $M-1$ predicted sketch control points and $H$ action steps form two token streams, with separate input projections into a shared embedding space. The denoiser is conditioned on $o_t$, $\mathbf q_t$, the optional task specification $\ell$, and the noise level of each stream. As shown in Figure~\ref{fig:method}a, a block-causal attention mask establishes directed sketch-to-action information flow. Sketch token $m\in\{1,\ldots,M-1\}$ attends only to sketch tokens with indices at most $m$. Action token $j\in\{1,\ldots,H\}$ attends to all sketch tokens and to action tokens with indices at most $j$. The sketch ordering follows the order of the B-spline basis supports along phase, while the action ordering follows execution time.

\textbf{Training with independent noise levels.} Following Diffusion Forcing~\citep{chenDiffusionForcing2024}, we sample the sketch and action noise coordinates $\gamma_Z$ and $\gamma_A$ independently and uniformly for each example; all tokens within a stream share its coordinate, while the Gaussian noise is sampled independently for each token (Figure~\ref{fig:method}b). We optimize
\begin{equation}
\label{eq:joint_loss}
\mathcal L_{\mathrm{PAM}}=\sum_{s\in\{Z,A\}}\lambda_s\,\mathbb E\bigl\lVert\mathbf U_s-f_\theta^{s}(\mathbf Z_t^{\gamma_Z},\mathbf A_t^{\gamma_A},\gamma_Z,\gamma_A,\mathbf c_t)\bigr\rVert^2,
\end{equation}
where $\mathbf U_s$ stacks the per-token targets of stream $s$, $f_\theta^{s}$ is the corresponding output head, the squared error is averaged over tokens and feature dimensions, and $\lambda_Z$ and $\lambda_A$ are loss weights.

\begin{wrapfigure}[12]{r}{0.45\textwidth}
  \vspace{\ifarxiv-1.6\else-0.5\fi\baselineskip}
  \setlength{\columnwidth}{\linewidth}
  \begin{algorithm}[H]
    \caption{Staggered denoising}
    \label{alg:sampling}
    \small
    \begin{algorithmic}[1]
      \Require $o_t,\mathbf q_t,\ell$; lead $\eta$; base steps $K$
      \State $\mathbf Z,\mathbf A\sim\mathcal N(\mathbf 0,\mathbf I)$
      \For{$k=0,\ldots,K+\lceil\eta K\rceil-1$}
        \State $b\leftarrow 1+\eta-k/K$;\quad $\gamma_Z,\gamma_A\leftarrow$ Eq.~\eqref{eq:front}
        \State $(\mathbf u_Z,\mathbf u_A)\leftarrow f_\theta(\mathbf Z,\mathbf A,o_t,\mathbf q_t,\ell,\gamma_Z,\gamma_A)$
        \State $\mathbf Z\leftarrow\textsc{Update}(\mathbf Z,\mathbf u_Z,\gamma_Z)$
        \State $\mathbf A\leftarrow\textsc{Update}(\mathbf A,\mathbf u_A,\gamma_A)$
      \EndFor
      \Ensure $\mathbf A$ for execution; $\mathbf Z$ for inspection only
    \end{algorithmic}
  \end{algorithm}
\end{wrapfigure}
\textbf{Staggered denoising.}
At inference, we initialize both streams with Gaussian noise and use a schedule that denoises the sketch ahead of the actions. We use a sketch lead $\eta\in[0,1]$ and assign the stream-wise schedules
\begin{equation}
\label{eq:front}
\gamma_Z=\operatorname{clip}(b-\eta,0,1),\quad
\gamma_A=\operatorname{clip}(b,0,1),
\end{equation}
where $b$ is a shared scheduling variable that starts at $1+\eta$ and decreases by $1/K$ until both coordinates reach zero, with $K$ the number of base sampling steps (Algorithm~\ref{alg:sampling}). A positive lead advances sketch denoising relative to action denoising, providing cleaner sketch inputs during action generation (Figure~\ref{fig:method}c), while $\eta=0$ assigns the same noise level to both streams at every step. At each step, $f_\theta$ reads both streams at their current noise levels, and \textsc{Update} applies one reverse step of the base sampler (DDPM, DDIM, or Euler) to each stream from its current to its next noise coordinate; it is the identity when the coordinate is unchanged. The sampler uses $K+\lceil\eta K\rceil$ NFE. When $\eta K$ is an integer, the first $\eta K$ iterations update only the sketch and the last $\eta K$ update only the actions. In this case, each stream receives exactly $K$ updates.

\section{Experiments}
\label{sec:experiments}
Our experiments address three questions. \textbf{(Q1)} Does PAM improve task performance over action-only policies in simulation and on real robots? \textbf{(Q2)} How do the sketch representation, directed attention, and denoising schedule contribute to performance? \textbf{(Q3)} To what extent does action generation depend on the predicted sketch?

\subsection{Setup}
\label{sec:setup}

\begin{figure}[t]
  \centering
  \includegraphics[width=\textwidth]{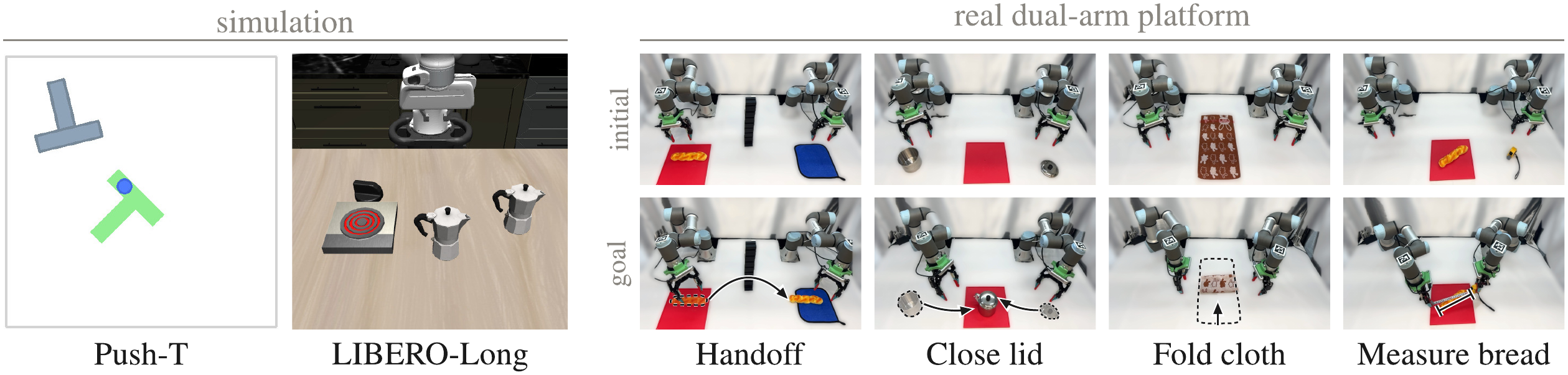}
  \caption{Overview of the simulation and real-world tasks. Push-T and LIBERO-Long in simulation, and four tasks on a real dual-arm platform shown at the initial state (top) and at the goal state (bottom). The Push-T goal is the green target pose; each LIBERO-Long task is specified by a language instruction (shown: \emph{put both moka pots on the stove}).}
  \label{fig:tasks}
\end{figure}

\noindent\textbf{Tasks.}
We evaluate PAM on two simulation benchmarks and four real-world bimanual tasks, shown in Figure~\ref{fig:tasks}. Push-T~\citep{chiDiffusionPolicy2023} requires pushing a T-shaped block to a target pose and supports evaluation with both state and image observations. LIBERO-Long~\citep{liu2023libero} comprises ten language-conditioned, long-horizon manipulation tasks on a simulated Franka arm. Our real-world tasks are \emph{handoff}, \emph{close lid}, \emph{fold cloth}, and \emph{measure}, performed on a dual-UR3 platform.

\noindent\textbf{Policies and baselines.}
We evaluate PAM-DP on Push-T and the real-world tasks, and PAM-VLA on LIBERO-Long. PAM-DP is trained from scratch using diffusion, with transformer-based Diffusion Policy~\citep{chiDiffusionPolicy2023} as its action-only baseline. PAM-VLA adds the sketch to the rectified-flow action expert of FLOWER~\citep{reussFLOWERDemocratizingGeneralist2025}, which serves as its pretrained action-only baseline. We also compare with B-spline Policy~\citep{hanBSplinePolicy2026}, which represents executable actions as spline curves. We compare with the transformer-based Diffusion Policy rather than its convolutional variant, because PAM represents the sketch and actions as token streams in a transformer.

\noindent\textbf{Implementation.}
Both PAM-DP and PAM-VLA independently sample sketch and action noise levels, with a shared level within each stream. We use a sketch lead of $\eta=1/8$ across all tasks unless otherwise specified. We use $M=12$ control points for Push-T, $M=16$ for LIBERO-Long, and $M=12$, $14$, $16$, and $14$ for handoff, close lid, fold cloth, and measure, respectively. In the real-world tasks, the successful evaluation episodes span roughly 250 to 370 control steps on average. These capacities satisfy the reconstruction criterion audited in Appendix~\ref{app:spec}. Appendix~\ref{app:arch} describes the model implementation, and Appendix~\ref{app:configs} summarizes training and evaluation settings.

\subsection{Simulation}
\label{sec:sim}
\begin{table}[t]
  \centering
  \begin{minipage}[t]{0.55\textwidth}
    \centering
    \caption{Push-T coverage with state and image observations. We report the maximum and the mean over the last 10 checkpoints.}
    \label{tab:pusht}
    \small
    \setlength{\tabcolsep}{3pt}
    \begin{tabularx}{\linewidth}{Xcc}
      \toprule
      Method & State & Image \\
      \midrule
      LSTM-GMM {\scriptsize\citep{mandlekarRobomimic2021}} & 0.67 / 0.61 & 0.69 / 0.54 \\
      IBC {\scriptsize\citep{florenceImplicitBehavioralCloning2022}} & 0.90 / 0.84 & 0.75 / 0.64 \\
      Diffusion Policy {\scriptsize\citep{chiDiffusionPolicy2023}} & 0.95 / 0.79 & 0.78 / 0.66 \\
      B-spline Policy {\scriptsize\citep{hanBSplinePolicy2026}} & 0.92 / 0.74 & 0.76 / 0.63 \\
      \midrule
      \rowcolor{black!7}
      PAM-DP (ours) & \textbf{0.96 / 0.88} & \textbf{0.82 / 0.72} \\
      \bottomrule
    \end{tabularx}
  \end{minipage}\hfill
  \begin{minipage}[t]{0.43\textwidth}
    \centering
    \caption{LIBERO-Long success rate. PAM-VLA results report mean~$\pm$~standard deviation over 5 training seeds.}
    \label{tab:libero}
    \small
    \setlength{\tabcolsep}{3pt}
    \begin{tabularx}{\linewidth}{Xc}
      \toprule
      Method & Success (\%) \\
      \midrule
      $\pi_0$ {\scriptsize\citep{blackVisionLanguageActionFlow2024}} & 85.2 \\
      $\pi_{0.5}$-KI {\scriptsize\citep{driessKnowledgeInsulating2025}} & 85.8 \\
      OpenVLA-OFT {\scriptsize\citep{kimOpenVLAOFT2025}} & 94.5 \\
      FLOWER {\scriptsize\citep{reussFLOWERDemocratizingGeneralist2025}} & 94.9 $\pm$ 1.2 \\
      \midrule
      \rowcolor{black!7}
      PAM-VLA (ours) & \textbf{95.2 $\pm$ 0.4} \\
      \bottomrule
    \end{tabularx}
  \end{minipage}
\end{table}

\noindent\textbf{Push-T.}
We evaluate PAM-DP with state and image observations using the dataset and evaluation protocol of \citet{chiDiffusionPolicy2023}. The configuration $\mathbf q_t$ is the measured two-dimensional pusher position, and the sketch represents its remaining path in both observation settings. The state-based setting allows us to examine the sketch without the additional challenge of visual perception. We measure performance by target coverage, the fraction of the target region covered by the block. Table~\ref{tab:pusht} reports the maximum over checkpoints and the mean over the last ten checkpoints, following \citet{chiDiffusionPolicy2023}, with 3 seeds and 50 initial conditions per evaluation.
PAM-DP achieves higher coverage than the action-only transformer and B-spline Policy in both observation settings. Compared with Diffusion Policy, the last-ten checkpoint average increases from $0.79$ to $0.88$ with state observations and from $0.66$ to $0.72$ with image observations. The corresponding maximum scores increase from $0.95$ to $0.96$ and from $0.78$ to $0.82$. The gains in the checkpoint average indicate that the differences are not limited to the best checkpoint.
B-spline Policy, which uses a spline to parameterize executable actions, does not improve over the dense transformer baseline in this setting, whereas using the spline to represent the remaining path alongside dense actions does.

\noindent\textbf{LIBERO-Long.}
We evaluate PAM-VLA on the ten language-conditioned tasks of LIBERO-Long following the protocol of \citet{reussFLOWERDemocratizingGeneralist2025}, with 50 trials per task, and report mean success rate and standard deviation over 5 training seeds. FLOWER also serves as the direct action-only baseline.
Table~\ref{tab:libero} shows that PAM-VLA increases mean success rate from FLOWER's $94.9\%$ to $95.2\%$ under the same evaluation protocol, with a standard deviation of $0.4$ percentage points across five training seeds. Since FLOWER already approaches the performance ceiling on LIBERO-Long, the remaining margin for improvement is small.

\subsection{Real-World Dual-Arm Manipulation}
\label{sec:real}
We evaluate PAM-DP on a dual-UR3 platform equipped with parallel grippers and three RGB cameras. The policy observes the camera images and robot joint positions, and outputs 14-dimensional joint and gripper commands at 8\,Hz. We collect 81, 81, 48, 89 demonstrations for handoff, close lid, fold cloth, and measure, respectively, with randomized initial object placements. In \emph{handoff}, the arms transfer a loaf of bread from the red board to the blue mat while clearing a barrier. In \emph{close lid}, one arm places a pot on the red plate while the other aligns and places its lid on the pot. The \emph{fold cloth} task requires coordinated corner grasps and successive folds, while \emph{measure} requires the arms to position and hold a tape measure along a loaf. We train and evaluate PAM-DP, transformer-based Diffusion Policy, and B-spline Policy on these tasks. Each method is evaluated over 20 episodes per task, with initial object placements sampled from the same distribution.

\begin{table}[t]
  \centering
  \caption{Real-world task performance. Success is reported as completed episodes out of 20 trials per task. Completion time is measured in seconds over successful episodes.
  }
  \label{tab:real}
  \footnotesize
  \begin{tabularx}{\textwidth}{ll*{5}{>{\centering\arraybackslash}X}}
    \toprule
    Method & Metric & Handoff & Close lid & Fold cloth & Measure & Overall \\
    \midrule
    Diffusion Policy & Success & 9/20 & 9/20 & 10/20 & 10/20 & 47.5\% \\
    & Time (s) & 36.2\,$\pm$\,4.6 & 44.6\,$\pm$\,9.2 & 41.9\,$\pm$\,1.3 & 42.8\,$\pm$\,3.6 & 41.4 \\
    \addlinespace
    B-spline Policy & Success & 8/20 & 8/20 & 10/20 & 9/20 & 43.8\% \\
    & Time (s) & 39.9\,$\pm$\,7.2 & 42.2\,$\pm$\,11.9 & 42.0\,$\pm$\,1.2 & 40.6\,$\pm$\,5.3 & 41.2 \\
    \midrule
    \rowcolor{black!7}
    PAM-DP (ours) & Success & \textbf{13/20} & \textbf{14/20} & \textbf{18/20} & \textbf{15/20} & \textbf{75.0\%} \\
    \rowcolor{black!7}
    & Time (s) & 35.8\,$\pm$\,8.0 & 45.3\,$\pm$\,11.1 & 42.4\,$\pm$\,2.4 & 45.2\,$\pm$\,6.4 & 42.3 \\
    \bottomrule
  \end{tabularx}
\end{table}

\begin{figure}[t]
  \centering
  \includegraphics[width=\textwidth]{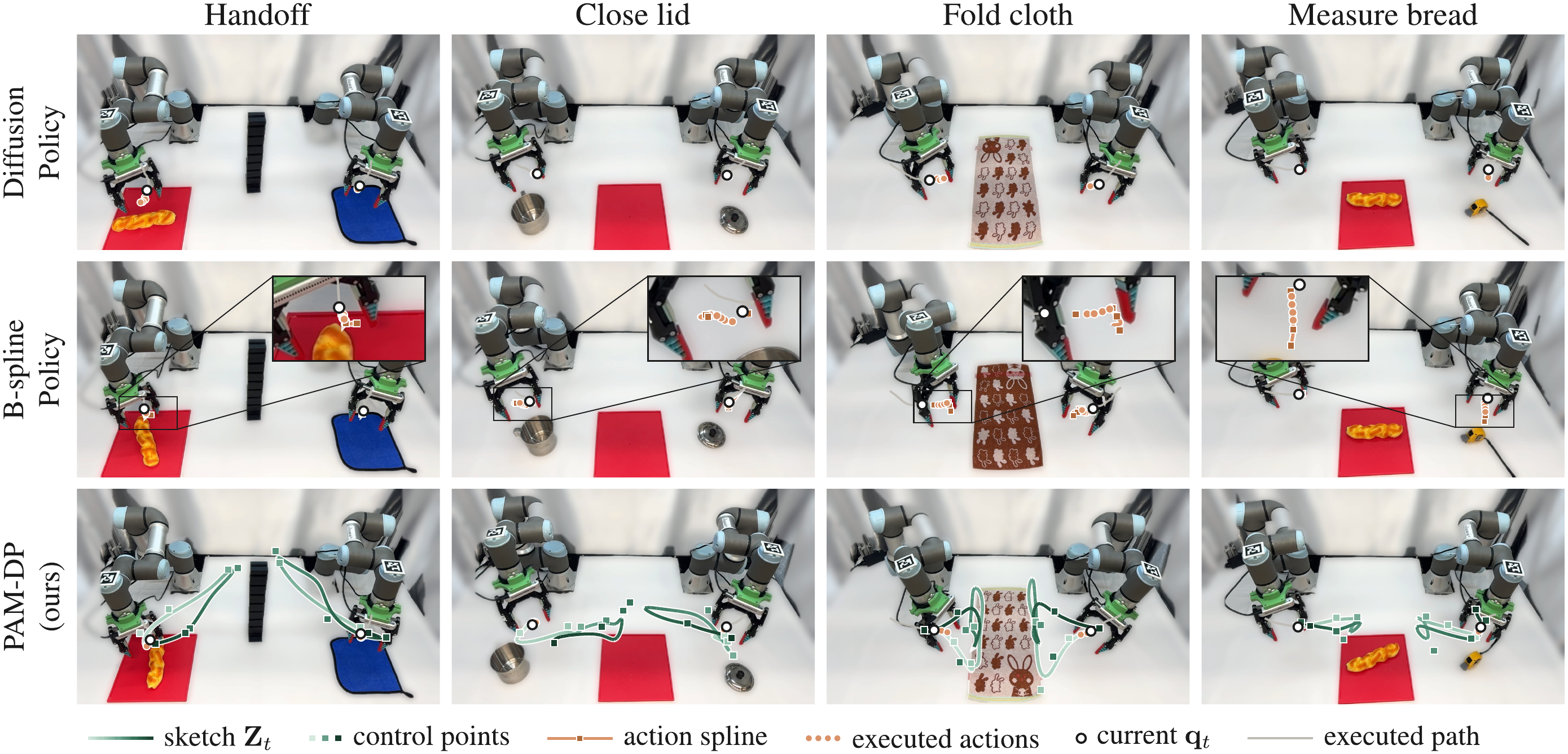}
  \caption{Policy predictions on the four real-world tasks, projected into the external camera view. For B-spline Policy, orange squares and curves show the control points and the action spline.
  }
  \label{fig:real_pred}
\end{figure}

\noindent\textbf{Task performance.}
Table~\ref{tab:real} reports success rates and completion times. PAM-DP succeeds in 60 of 80 episodes ($75.0\%$), compared with 38 ($47.5\%$) for Diffusion Policy and 35 ($43.8\%$) for B-spline Policy (two-sided Fisher exact test, $p<0.001$ for both). The improvement holds on all four tasks despite their different objects and coordination demands. B-spline Policy does not exceed Diffusion Policy on any task, suggesting that the gain comes from representing the remaining path rather than from spline parameterization itself. PAM-DP takes slightly longer on successful episodes, consistent with the additional NFE from the sketch lead and the extra sketch tokens processed at each evaluation.

\noindent\textbf{Predicted motion.}
Figure~\ref{fig:real_pred} visualizes policy predictions early in task execution. The recording camera is separate from the three policy cameras and is used only for visualization. We map predicted joint configurations to tool positions through forward kinematics and project them into the recording view.
Diffusion Policy directly generates a short-horizon action chunk, while B-spline Policy predicts control points that define an executable spline from which the actions are sampled.
The action predictions describe upcoming commands, while PAM-DP's sketch extends over the remaining motion. In \emph{handoff}, for example, the sketch paths extend toward the transfer region near the barrier, whereas the action chunks remain near the current gripper positions.

\begin{figure}[t]
  \centering
  \includegraphics[width=\textwidth]{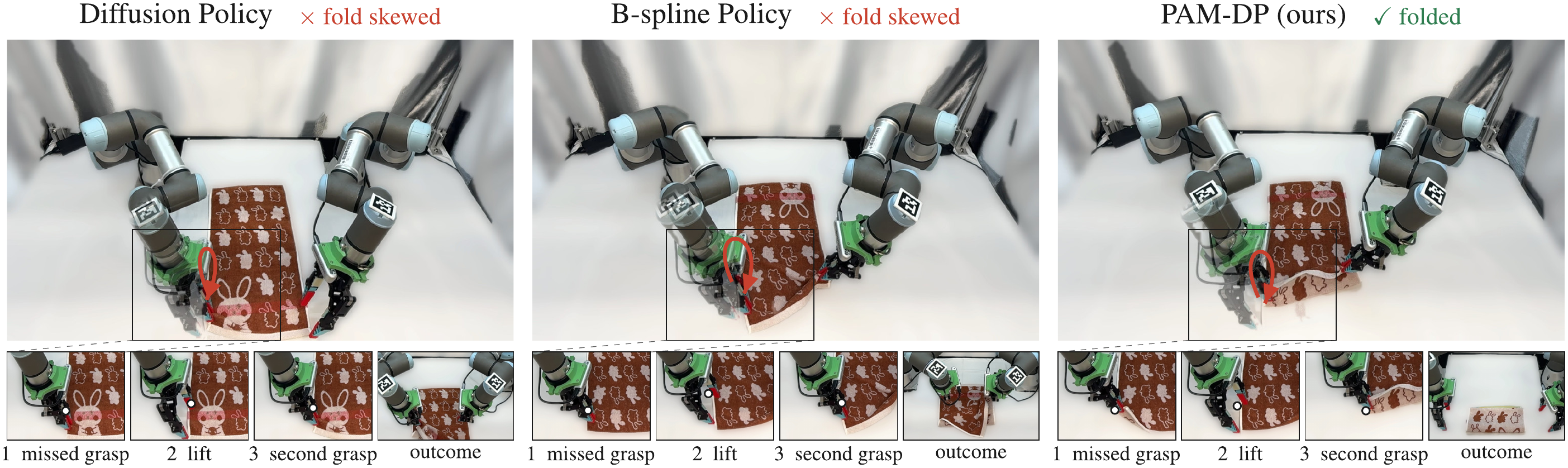}
  \caption{Grasp retries on \emph{fold cloth}. Each column shows an episode in which the first grasp missed. The large image shows the scene after the second grasp, with a translucent overlay marking the gripper at the top of its lift and an arrow indicating the lift and re-descent.
  }
  \label{fig:real_fail}
\end{figure}

\begin{wraptable}[9]{r}{0.45\textwidth}
  \vspace{-0.5\baselineskip}
  \centering
  \caption{Failed episodes across 80 real-world trials per method.}
  \label{tab:real_fail}
  \footnotesize
  \setlength{\tabcolsep}{4pt}
  \begin{tabular}{lcccc}
    \toprule
    Method & Total & Miss & Slip & Coord. \\
    \midrule
    Diffusion Policy & 42 & 22 & 12 & 8 \\
    B-spline Policy & 45 & 28 & 15 & 2 \\
    \rowcolor{black!7}
    PAM-DP (ours) & \textbf{20} & \textbf{10} & \textbf{9} & \textbf{1} \\
    \bottomrule
  \end{tabular}
\end{wraptable}
\noindent\textbf{Failure analysis.}
Table~\ref{tab:real_fail} groups failed episodes into missed grasps, object loss after grasping, and coordination errors between the arms. Coordination errors include transfer misalignment and collisions with the barrier. Each failed episode receives one label corresponding to the error that directly caused task failure. When errors occur in sequence, we record the earliest error that led to failure. Errors recovered within an ultimately successful episode are not included in these counts.

Compared with Diffusion Policy, PAM-DP has fewer failed episodes attributed to missed grasps, decreasing from 22 to 10, and coordination errors, decreasing from 8 to 1. Failures attributed to object loss decrease from 12 to 9. Figure~\ref{fig:real_fail} provides qualitative examples of grasp retries on \emph{fold cloth}. Following an unsuccessful initial grasp, all three policies lift the gripper and re-approach the cloth. In the displayed PAM-DP episode, the second grasp reaches the cloth corner and the fold completes. The displayed Diffusion Policy and B-spline Policy episodes instead grasp away from the corner and produce skewed folds. These examples illustrate individual recovery attempts rather than a comparison of overall recovery success rates.

\subsection{Analysis}
\label{sec:analysis}
\newcommand{\ablationintro}{\textbf{Component ablations.} We study the sketch parameterization, attention structure, and training noise distribution on image-based Push-T (Table~\ref{tab:pusht_ablation}). All variants are derived from the same PAM-DP configuration and use training seed 42. Except for the specified ablation, we hold the backbone, conditioning, data, optimizer, training budget, and evaluation episodes fixed. We use 100 DDPM sampling steps with zero sketch lead to isolate representation and training effects from staggered denoising and its additional NFE. We report mean coverage and standard deviation across the last ten evaluated checkpoints, each evaluated on the same 50 initial conditions.}
\ifarxiv
\ablationintro

\fi
\begin{wraptable}[12]{r}{0.4\textwidth}
  \vspace{\ifarxiv0\else-0.5\fi\baselineskip}
  \centering
  \caption{Component ablations on image-based Push-T.}
  \label{tab:pusht_ablation}
  \footnotesize
  \setlength{\tabcolsep}{3pt}
  \begin{tabularx}{\linewidth}{Xc}
    \toprule
    Variant & Coverage \\
    \midrule
    \rowcolor{black!7}
    PAM-DP, zero lead & $\mathbf{0.79} \pm 0.02$ \\
    \midrule
    Action-only backbone & 0.69 $\pm$ 0.07 \\
    Auxiliary sketch only & 0.75 $\pm$ 0.04 \\
    Time-indexed sketch & 0.74 $\pm$ 0.04 \\
    Bidirectional attention & 0.72 $\pm$ 0.06 \\
    Shared-time training & 0.74 $\pm$ 0.07 \\
    \bottomrule
  \end{tabularx}
\end{wraptable}
\ifarxiv\else
\ablationintro

\fi
The \emph{action-only backbone} removes the sketch tokens and sketch loss from PAM-DP. The \emph{auxiliary-only} variant retains sketch prediction and supervision but removes attention from action queries to sketch keys. Sketch supervision can therefore affect actions through shared parameters, but actions cannot read the sketch tokens. The \emph{time-indexed sketch} variant replaces uniform arc-length resampling with uniform-time resampling before B-spline fitting, retaining the same control-point count, token budget, loss weights, and attention structure. The \emph{bidirectional} variant allows all target tokens to attend to one another and to the complete observation memory. The \emph{shared-time} variant ties the sketch and action noise levels during training, whereas zero lead ties them only during sampling.
PAM-DP with zero lead achieves mean coverage of $0.79$, compared with $0.69$ for the action-only backbone and $0.75$ for auxiliary-only prediction. Time-indexed sketches, bidirectional attention, and shared-time training reduce coverage to $0.74$, $0.72$, and $0.74$, respectively. Each component therefore contributes to performance, even without staggered denoising.

\begin{wraptable}[11]{r}{0.45\textwidth}
  \vspace{-1\baselineskip}
  \centering
  \caption{Sketch lead ablation on Push-T with a 16-step DDIM base budget. }
  \label{tab:pusht_lead}
  \footnotesize
  \setlength{\tabcolsep}{1.75pt}
  \begin{tabularx}{\linewidth}{Xcccc}
    \toprule
    \mbox{Lead $\eta$} & Offset & NFE & Coverage & Delay (ms) \\
    \midrule
    0     & 0  & 16 & $0.779 \pm 0.015$ & 153.0 \\
    1/16  & 1  & 17 & $0.801 \pm 0.018$ & 155.2 \\
    \rowcolor{black!7}
    1/8   & 2  & 18 & $0.805 \pm 0.011$ & 158.1 \\
    1/4   & 4  & 20 & $0.785 \pm 0.029$ & 193.1 \\
    1/2   & 8  & 24 & $0.788 \pm 0.023$ & 230.0 \\
    1     & 16 & 32 & $\mathbf{0.809} \pm 0.027$ & 280.0 \\
    \bottomrule
  \end{tabularx}
\end{wraptable}
Sketch-lead sensitivity. We vary the sketch lead with a 16-step DDIM base budget (Table~\ref{tab:pusht_lead}). We report mean coverage and standard deviation over three sampler-noise seeds, each evaluated on the same 50 initial conditions. All positive leads yield higher mean coverage than zero lead. The default $\eta=1/8$, highlighted by shading, increases coverage from $0.779$ to $0.805$ with 18 rather than 16 NFE. Increasing the lead to $\eta=1$ yields $0.809$ coverage but requires 32 NFE. These results support a modest lead as a practical trade-off between coverage and computation. Appendix~\ref{app:evaluation} details the lead-evaluation protocol.

\begin{figure}[t]
  \centering
  \includegraphics[width=\textwidth]{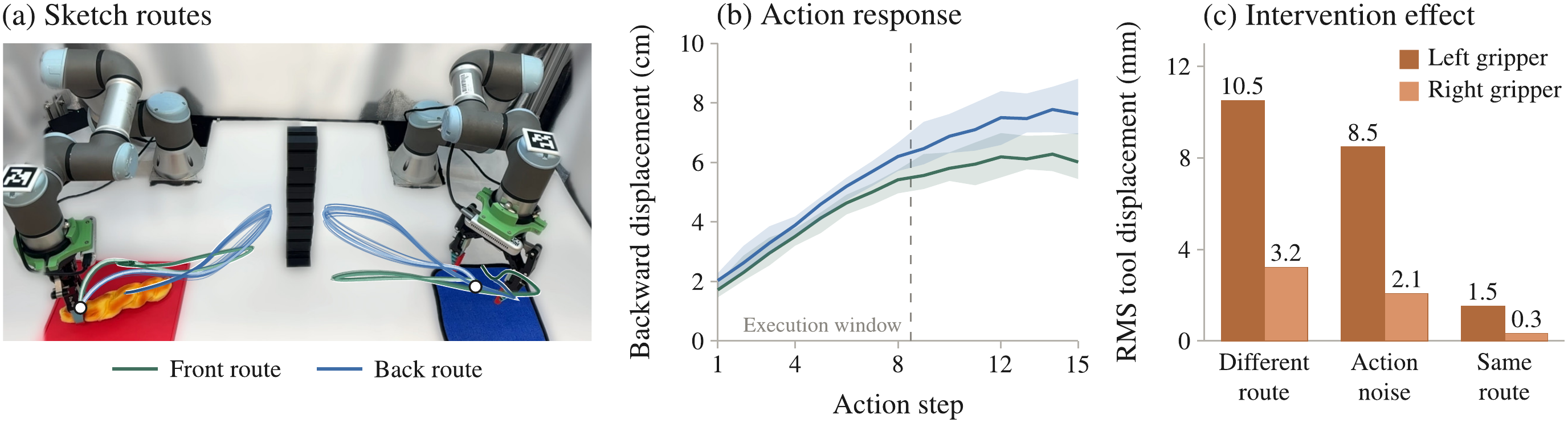}
  \caption{Offline sketch intervention. (a) Alternative sketch routes. (b) Predicted left-gripper motion. (c) Action sensitivity to sketch and noise changes.}
  \label{fig:intervention}
\end{figure}
\noindent\textbf{Sketch intervention.} We test sketch dependence offline at a fixed handoff observation using routes passing in front of or behind the barrier (Figure~\ref{fig:intervention}a). We hold the observation and initial action noise fixed and replace the sketch denoising trajectory at matching noise levels. Back-route sketches induce greater predicted backward motion of the left gripper (Figure~\ref{fig:intervention}b). At action step 8, different-route replacement changes the predicted left-tool position by 10.5\,mm RMS, compared with 8.5\,mm for action-noise resampling and 1.5\,mm for same-route replacement (Figure~\ref{fig:intervention}c). Changing the sketch route alters the predicted motion even when the observation and action noise remain fixed. Appendix~\ref{app:intervention} describes donor selection, noise pairing, and the displacement metric.

\section{Conclusion}
\label{sec:conclusion}
We introduced Proprioceptive Action Models, which jointly generate a timing-free B-spline sketch of the remaining configuration-space path and a short executable action chunk through directed attention and staggered denoising. Adding the sketch improves over the corresponding action-only policies on Push-T, LIBERO-Long, and four real-world bimanual tasks. Several limitations remain. The sketch represents the robot's motion without modeling object dynamics or contact. Staggered denoising requires additional NFE, and our lead sweep does not separate the schedule from this added computation. Future work will explore sketches that capture object interactions and evaluate denoising schedules under matched inference budgets.

%% file: content/appendix.tex
\section{Sketch Construction and Theoretical Properties}
\label{app:sketch_construction}
\label{app:spec}

\subsection{Path Preprocessing and Phase Parameterization}
\label{app:path_preprocessing}
At each timestep $t$, we use the remaining measured configuration path $\mathbf q_{t:T}$ to construct the sketch target. For the dual-arm platform, we concatenate the two arms' joint positions and exclude gripper states. We subtract $\mathbf q_t$ and compute cumulative Euclidean arc length with equal weight across configuration dimensions, following Eq.~\ref{eq:phase}.

Gripper commands remain in the action stream. Excluding grippers from the sketch prevents their discrete state changes from affecting the arm-motion phase. We linearly interpolate the relative path onto $N=64$ uniformly spaced phase samples, including both endpoints. Before interpolation, we retain the initial sample and each subsequent sample whose arc-length increment from the preceding original sample exceeds a numerical tolerance $\epsilon$. After interpolation, we explicitly set the first and last positions to $\mathbf 0$ and $\mathbf q_T-\mathbf q_t$. The resulting samples are used for the spline fit.

We set $\epsilon=\max(10^{-8}R,\epsilon_{\mathrm{mach}})$, where $R$ is the largest per-coordinate range over the full configuration array supplied to target construction and $\epsilon_{\mathrm{mach}}$ is double-precision machine epsilon. In the dual-arm implementation, this array contains the configurations from the entire replay buffer. If the remaining path has a single sample or total length $L<\epsilon$, we set its sketch target to zero. The tolerance $\epsilon$ removes numerically repeated samples but does not threshold measurement noise; joint-encoder jitter during stationary intervals therefore contributes small phase increments.

\subsection{B-spline Representation and Target Fitting}
\label{app:sketch_fitting}
We fit the resampled relative path with a clamped cubic B-spline using $M$ control points. The knot vector contains four repeated knots at each endpoint, zero and one, with interior knots at $j/(M-3)$ for $j=1,\ldots,M-4$. We use the standard Cox--de Boor basis $B_{m,3}$~\citep{deBoor2001}.

Let $\Delta\mathbf q_{t,n}$ denote the resampled relative configuration at phase $\bar\phi_n=n/(N-1)$ for $n=0,\ldots,N-1$. We fit the coefficients by endpoint-constrained least squares,
\begin{equation}
\begin{aligned}
\mathbf Z_t^\star
&=\underset{\mathbf Z_t}{\arg\min}\;
\sum_{n=0}^{N-1}
\left\|\sum_{m=0}^{M-1}B_{m,3}(\bar\phi_n)\,\mathbf z_{t,m}
-\Delta\mathbf q_{t,n}\right\|_2^2,\\
&\text{subject to}\quad
\mathbf z_{t,0}=\mathbf 0,\qquad
\mathbf z_{t,M-1}=\mathbf q_T-\mathbf q_t.
\end{aligned}
\end{equation}
We subtract the fixed endpoint contribution and solve for the interior coefficients using a pseudoinverse. The basis matrix and pseudoinverse are precomputed for each choice of $M$ and $N$.

\paragraph{Temporal invariance.}
\label{app:sketch_invariance}
\label{app:proofs}
For a Lipschitz path $\mathbf q:[0,D]\to\mathbb R^{d_q}$ of positive arc length and an absolutely continuous, non-decreasing, surjective time map $\rho:[0,D']\to[0,D]$, cumulative arc length satisfies $s_{\mathbf q\circ\rho}=s_{\mathbf q}\circ\rho$. The normalized arc-length path is therefore unchanged, so a fixed phase grid, basis, and coefficient budget $M$ yield identical fitting targets, endpoint constraints, and pseudoinverse solutions. This invariance includes changes in speed and inserted pauses that preserve traversal order, but not reversal or additional retracing. Finite sampling, measurement noise, and numerical tolerances can still produce differences between separately recorded demonstrations. Removing timing holds for every fixed $M$, whereas geometric fidelity depends on the capacity selected below. For the dual-arm platform, the invariance applies to a common reparameterization of both arms. Changing the timing of one arm relative to the other alters the path in the joint configuration space, so the sketch retains the relative coordination of the two arms.

\subsection{Capacity Audit and Normalization}
\label{app:sketch_capacity}
\label{app:sketch_normalization}
We audit the adopted coefficient budgets after training using a common reconstruction metric. We measure $E_{\mathrm{rec}}(M)$ as the scalar RMSE between each fitted path and its 64 uniformly spaced phase targets, weighting anchors, phase samples, and configuration dimensions equally. All capacity audits use training episodes. Push-T and the dual-arm tasks use training-sampler anchors, while LIBERO uses every frame. We compute the per-step motion reference $\delta_{\mathrm{step}}$ as the root mean square of configuration increments over all adjacent frame pairs within training episodes and all configuration dimensions, retaining stationary steps. LIBERO recordings are sampled at 20\,Hz.

\begin{figure}[htbp]
  \centering
  \includegraphics[width=\linewidth]{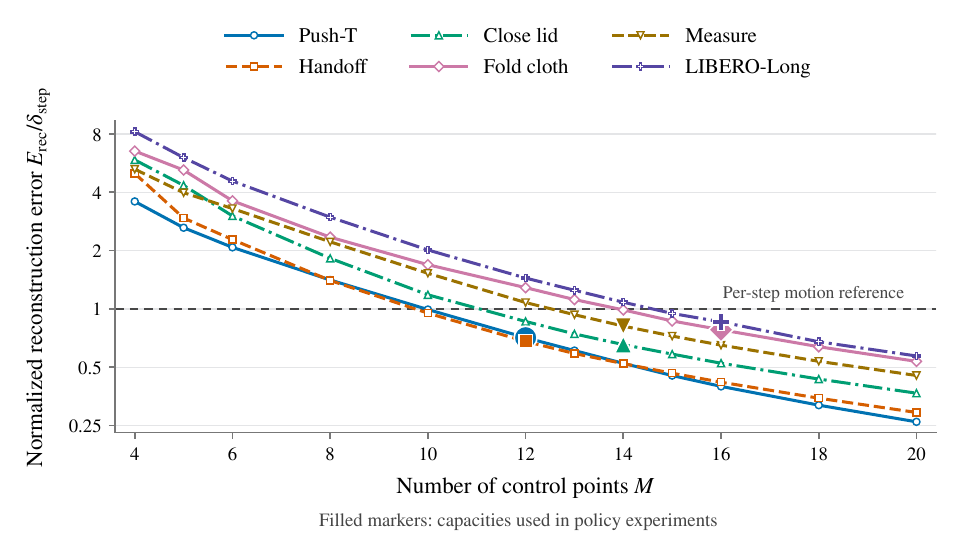}
  \caption{Post-hoc sketch capacity audit on training anchors. Reconstruction RMSE is normalized by the task-specific per-step motion reference and shown on a logarithmic vertical axis. The dashed line marks a ratio of one; filled markers indicate the capacities used in the policy experiments. The reference values are $8.04$ pixels for Push-T and $0.870$, $0.700$, $0.765$, $0.660$, and $0.665$ degrees for handoff, close lid, fold cloth, measure, and LIBERO-Long, respectively.}
  \label{fig:sketch_capacity_audit}
\end{figure}

Figure~\ref{fig:sketch_capacity_audit} shows reconstruction errors normalized by each task's per-step motion reference across the common candidate set and marks the capacities used in the policy experiments. The adopted capacities satisfy $E_{\mathrm{rec}}(M)<\delta_{\mathrm{step}}$ on all six tasks; we do not require the smallest budget satisfying this criterion. For LIBERO-Long, the per-step reference is $0.0116$\,rad and the reconstruction RMSE at $M=16$ is $0.0098$\,rad, giving a normalized error of approximately $0.845$. These post-hoc audits verify the reconstruction quality of the capacities used in training.

We normalize each predicted coefficient and configuration dimension independently using training-set minima and maxima. In the dual-arm implementation, these statistics use targets indexed by the training sampler; nonconstant dimensions map to $[-1,1]$, while nearly constant dimensions retain unit scale and only subtract the training minimum. Separately from the training-set capacity audit, we audit clipping on fitted validation targets by normalizing their coefficients, clamping to $[-1,1]$, inverting the transform, and decoding the paths. Scalar reconstruction RMSE changes from $0.568$ to $0.571$ degrees for handoff, $0.528$ to $0.541$ for close lid, $0.563$ to $0.566$ for fold cloth, and $0.596$ to $0.602$ for measure. Push-T remains at $3.08$ pixels at the reported precision. These values measure distortion of fitted targets rather than model prediction error.

\section{Model Implementation}
\label{app:arch}
This section specifies token processing, training targets, and sampling updates for the joint policy in Section~\ref{sec:joint}. Experiment-specific dimensions, optimization settings, and inference budgets are collected in Appendix~\ref{app:configs}.

\subsection{Architecture}
\label{app:model_architecture}
\textbf{Token streams.} PAM-DP projects each of the $M-1$ normalized sketch coefficients and each of the $H$ normalized action steps into a shared embedding space using separate linear layers. Each stream has learned positional embeddings, and sketch tokens receive an additional learned type embedding. The streams are concatenated, processed by a shared transformer decoder, and mapped back to their respective dimensions by separate output heads.

\textbf{Attention and observations.} Sketch tokens attend causally within the sketch stream and never read action tokens. Action tokens attend to all sketch tokens and causally within the action stream.

The observation encoder processes each timestep independently, combining image and low-dimensional inputs into a feature vector. These features are projected, combined with learned positional embeddings, and passed through a position-wise MLP to form the cross-attention memory. There is no temporal mixing in this conditioning network. Sketch queries read all observation positions, whereas action queries read observation positions up to their own time index, following the transformer Diffusion Policy.

\textbf{Noise conditioning.} The main Push-T model uses adaLN-zero conditioning. The sketch and action timesteps are embedded separately and broadcast to the tokens of their respective streams. Each decoder block uses these embeddings to modulate the normalized self-attention and feed-forward inputs and to gate the self-attention, observation cross-attention, and feed-forward residual updates. The modulation projections and output heads are zero-initialized. The final normalization is also time-conditioned.

\textbf{PAM-VLA.} PAM-VLA augments FLOWER's rectified-flow action expert with sketch prediction. Separate MLPs encode sketch coefficients and actions into a shared DiT, and separate linear heads predict their velocities. Rotary positional indices restart at zero for each stream. Sketch self-attention is causal and excludes action keys; actions attend to all sketch tokens and causally to other actions. Projected VLM features provide cross-attention context. Noise-level and control-frequency embeddings condition the tokens and adaptive normalization. The pretrained Florence-2 backbone, including its vision tower, is fine-tuned rather than frozen.

\subsection{Training and Sampling}
\label{app:model_training_sampling}
\textbf{Diffusion training.} The Push-T model uses a 100-step cosine diffusion schedule. For each example, it independently samples one discrete timestep for the sketch and another for the actions, uniformly over the training schedule. Every token in a stream shares its timestep, while Gaussian noise is sampled independently for every token and feature. In Eq.~\ref{eq:corruption}, the diffusion scales are $\alpha=\sqrt{\bar\alpha_k}$ and $\sigma=\sqrt{1-\bar\alpha_k}$ at scheduler index $k$, where $\bar\alpha_k$ is the cumulative signal-retention product. Here, the prediction target is the sampled noise $\boldsymbol\epsilon$, and $f_\theta$ estimates this target as $\hat{\boldsymbol\epsilon}$. Each stream's mean squared error is averaged over the batch, tokens, and feature dimensions before applying the weights in Eq.~\ref{eq:joint_loss}.

The real-world experiments use adaLN-zero conditioning and a 100-step cosine diffusion schedule with noise prediction. The independent-time model samples separate sketch and action timesteps, shared within each stream, while the shared-time control uses one timestep across both streams. Offline checkpoint validation uses synchronous 16-step DDIM. For the main real-world results, inference uses a sketch lead of $\eta=1/8$.

\textbf{Rectified-flow training.} PAM-VLA independently samples one uniform noise coordinate for the sketch stream and one for the action stream, shared by all tokens within the respective stream. It uses linear interpolation with $\alpha(\gamma)=1-\gamma$ and $\sigma(\gamma)=\gamma$, where zero is clean and one is noise. The velocity target is $\boldsymbol\epsilon_i-\mathbf x_i^0$, and each stream's squared velocity error is averaged before weighting.

\textbf{Discrete sketch lead.} The Push-T sampler realizes Algorithm~\ref{alg:sampling} as an offset in the inference timestep list, where the normalized coordinates $\gamma_Z$ and $\gamma_A$ index a list of $K$ scheduler timesteps and \textsc{Update} is one DDPM or DDIM step. The requested lead $\eta$ is converted to an integer offset $k_{\mathrm{lead}}=\lfloor\eta K+0.5\rfloor\in\{0,\ldots,K\}$, rounding half-integer offsets upward. For discrete sampling, Algorithm~\ref{alg:sampling} uses $k_{\mathrm{lead}}/K$ in place of $\eta$. For $K=100$ and $\eta=1/8$, the offset is 13, giving 113 NFE and exactly 100 updates per stream. Both streams start from independent Gaussian noise. The sampler first performs $k_{\mathrm{lead}}$ sketch updates while holding the action noise fixed. It then performs $K$ action updates, with the sketch advancing through its own schedule $k_{\mathrm{lead}}$ positions ahead. Once the sketch has completed all $K$ updates, it remains fixed while action denoising finishes. Each model call reads both streams at their current local timesteps, and each active stream uses its own predicted noise $\mathbf u$ in the update. This schedule requires $K+k_{\mathrm{lead}}$ NFE. Zero lead updates both streams synchronously; a lead of $K$ completes sketch generation before updating actions. We use $\eta=1/8$ for the main results. Once the sketch is fixed, its conditioning timestep remains the final index in the scheduler's inference list.

The Push-T training configuration evaluates checkpoints with 100-step DDPM and a sketch lead of $\eta=1/8$. Subsequent lead sweeps use deterministic DDIM with $\eta_{\mathrm{DDIM}}=0$, retaining the same 100-step training noise schedule. The main-result sketch lead is applied at inference; it does not change the training corruption distribution. Per-experiment samplers and base budgets are reported in Appendix~\ref{app:configs}.

\textbf{Rectified-flow sampling.} PAM-VLA runs Algorithm~\ref{alg:sampling} with \textsc{Update} as an explicit Euler step: token $i$ is updated as $\mathbf x_i\leftarrow\mathbf x_i-(\gamma_i^{\mathrm{old}}-\gamma_i^{\mathrm{new}})\mathbf u_i$, where $\mathbf u_i$ is the predicted velocity, and tokens whose noise level is unchanged remain fixed. The stream-wise noise coordinates follow Eq.~\ref{eq:front} directly, without discrete lead rounding. LIBERO inference uses $K=8$ and $\eta=1/8$, giving nine NFE. Final normalized sketch and action outputs are clipped to $[-1,1]$.

\textbf{Action execution and ablations.} After sampling, the policy inverts the sketch and action normalizers and restores the sketch's zero anchor. Push-T uses two observation steps and predicts 16 action positions, executing eight actions before replanning. The sketch is decoded for inspection but is never executed.

Shared-time training ties the two corruption levels during optimization. Zero-lead sampling instead ties their schedules only at inference and can be applied to an independently trained model. These are distinct ablations.

\section{Experimental Configurations and Evaluation Protocols}
\label{app:configs}

\subsection{Training and Inference Configurations}
\label{app:training_configs}
Table~\ref{tab:training_configs} summarizes the PAM-DP training settings. Push-T uses the image-based $H=16$ model and adopts the transformer hyperparameters reported by \citet{chiDiffusionPolicy2023}. The real-world settings are shared by handoff, close lid, fold cloth, and measure. The sketch loss weight is $0.15$ for close lid and $0.2$ for the other three real-world tasks. The action loss weight is one.

\begin{table}[t]
  \centering
  \caption{PAM-DP training settings. Real-world settings are shared by the four dual-arm tasks. Both configurations use EMA and a 100-step cosine diffusion schedule with noise prediction.}
  \label{tab:training_configs}
  \small
  \begin{tabular}{lcc}
    \toprule
    Setting & Push-T image & Real-world \\
    \midrule
    Decoder layers / heads / width & 8 / 4 / 256 & 8 / 4 / 256 \\
    Conditioning & adaLN-zero & adaLN-zero \\
    Attention dropout & 0.01 & 0 \\
    Observation / prediction / execution steps & 2 / 16 / 8 & 2 / 16 / 8 \\
    Image crop & $84\times84$ & $84\times84$ \\
    Batch size & 64 & 64 \\
    Optimizer & AdamW & AdamW \\
    Learning rate & $10^{-4}$ & $10^{-4}$ \\
    Adam coefficients & $0.9,\ 0.95$ & $0.9,\ 0.95$ \\
    Denoiser weight decay & $10^{-1}$ & $10^{-3}$ \\
    Observation-encoder weight decay & $10^{-6}$ & $10^{-6}$ \\
    Learning-rate schedule & Cosine & Cosine \\
    Warmup updates & 1,000 & 500 \\
    Training budget & 3,050 epochs & Matched across methods \\
    \bottomrule
  \end{tabular}
\end{table}

Table~\ref{tab:pam_vla_configs} reports the PAM-VLA settings. Florence-2-large, including its vision tower, is fully fine-tuned. The learning rate rises from $2\times10^{-6}$ to $2\times10^{-5}$ over 1,500 updates, remains constant for 3,000 updates, and follows cosine decay to $10^{-5}$ over 25,500 updates. It remains at $10^{-5}$ thereafter. EMA uses an inverse-power warmup with power $2/3$, inverse scale one, and maximum decay $0.9999$; evaluation uses the EMA weights.

\begin{table}[t]
  \centering
  \caption{PAM-VLA configuration on LIBERO-Long. The sketch has $M=16$ control points and predicts 15 coefficient vectors after fixing the first to zero.}
  \label{tab:pam_vla_configs}
  \small
  \begin{tabular}{lc}
    \toprule
    Setting & PAM-VLA \\
    \midrule
    Vision-language backbone & Florence-2-large \\
    Decoder layers / heads / width & 18 / 16 / 1024 \\
    Positional encoding & RoPE \\
    Attention / residual / MLP dropout & 0.1 / 0.1 / 0.1 \\
    Observation / prediction / execution steps & 1 / 10 / 10 \\
    Batch size & 16 \\
    Optimizer & AdamW \\
    Peak learning rate & $2\times10^{-5}$ \\
    Adam coefficients & $0.9,\ 0.95$ \\
    Weight decay & 0.05 \\
    Training limit & 45 epochs, at most 1,000 batches per epoch \\
    Precision & BF16 mixed precision \\
    Sketch / action loss weights & 0.2 / 1.0 \\
    Sampler / base steps / lead / NFE & Euler / 8 / $1/8$ / 9 \\
    \bottomrule
  \end{tabular}
\end{table}

PAM-DP also uses inference lead $1/8$ for the main results. The 16-step DDIM lead experiment has $k_{\mathrm{lead}}=2$ and therefore uses $K+k_{\mathrm{lead}}=18$ NFE at this lead.

\textbf{Baselines.} The direct comparisons use action-only Diffusion Policy for PAM-DP and FLOWER for PAM-VLA. B-spline Policy predicts executable action curves. On Push-T, B-spline Policy is trained with the transformer backbone for three training seeds and 3,050 epochs under the same protocol as Diffusion Policy. In the real-world experiments, all three methods use the same visual encoder, EMA, and number of optimizer updates on each task.

\subsection{Datasets and Evaluation Protocols}
\label{app:evaluation}
\textbf{Demonstrations.} Table~\ref{tab:dataset_splits} gives the available demonstrations and the seed-42 training/validation splits. Push-T follows the data split of \citet{chiDiffusionPolicy2023}. LIBERO uses all 500 demonstrations for training. Validation episodes are excluded from training-normalizer statistics.

\begin{table}[htbp]
  \centering
  \caption{Demonstration counts and seed-42 splits. Push-T is not listed and follows the split of \citet{chiDiffusionPolicy2023}. LIBERO has no demonstration-level validation split.}
  \label{tab:dataset_splits}
  \small
  \begin{tabular}{lrrr}
    \toprule
    Task & Available & Training & Validation \\
    \midrule
    LIBERO-Long & 500 & 500 & 0 \\
    Handoff & 81 & 73 & 8 \\
    Close lid & 81 & 73 & 8 \\
    Fold cloth & 48 & 41 & 7 \\
    Measure & 89 & 80 & 9 \\
    \bottomrule
  \end{tabular}
\end{table}

\textbf{Push-T.} Coverage is the maximum target overlap achieved during an episode, averaged over evaluation initial conditions. Table~\ref{tab:pusht} reports checkpoint aggregates over 50 initial conditions and three training seeds. For image-based training, the evaluation launcher assigns 50 consecutive initial-condition seeds starting at $100000(s+1)$ for training seed $s$. Following \citet{chiDiffusionPolicy2023}, we report the maximum coverage over checkpoints and the mean over the last ten checkpoints, with checkpoints evaluated every 50 epochs. Results for LSTM-GMM, IBC, and Diffusion Policy in Table~\ref{tab:pusht} are taken from \citet{chiDiffusionPolicy2023}; we train B-spline Policy and PAM-DP under the same protocol and report the same two aggregates.

The component ablations in Table~\ref{tab:pusht_ablation} use training seed 42, the same 50 initial conditions, 100-step DDPM, and zero lead. Their mean and standard deviation describe the last ten evaluated checkpoints, not independent training runs.

\textbf{Lead evaluation.} Table~\ref{tab:pusht_lead} uses a single seed-42 final-epoch checkpoint, without score-based checkpoint selection, and deterministic 16-step DDIM reconstructed from the training schedule. Each lead is evaluated on initial-condition seeds $4300000$--$4300049$ with three sampler-noise seeds. Initial noise is paired by episode and decision step across leads; reported standard deviations are across sampler-noise seeds. We report computational cost as NFE per action chunk.

\textbf{LIBERO-Long.} We evaluate each of the ten tasks over 50 trials per policy and report the mean success rate across tasks. The main table aggregates PAM-VLA over five training seeds. Baseline results in Table~\ref{tab:libero}, including FLOWER, are taken from their original publications, except for $\pi_0$, whose LIBERO-Long result is taken from \citet{kimOpenVLAOFT2025}.

\textbf{Real-world tasks.} Each method is evaluated over 20 episodes per task with initial object placements sampled from the same distribution. Each episode runs for at most 120\,s without human intervention; the policy may regrasp autonomously within this limit, and an episode that has not succeeded by 120\,s is counted as a failure. In \emph{handoff}, one arm grasps the loaf and passes it to the other arm around the barrier, which places it on the blue mat. In \emph{close lid}, the lid rests on the pot and the pot rests on the red plate; the lid need not be fully seated. In \emph{fold cloth}, the cloth is folded twice into the goal configuration shown in Figure~\ref{fig:tasks}. In \emph{measure}, the arms extend the tape measure along the loaf and then release it. Completion time is reported as mean $\pm$ standard deviation over successful episodes only; the overall time pools successful episodes across tasks. Each failed episode receives one label for the earliest error that led to failure, and recovered errors in successful episodes are not counted in Table~\ref{tab:real_fail}.

For the real-world tasks, offline checkpoint selection uses EMA predictions and episode-averaged physical joint-action RMSE, averaged over five inference seeds with synchronous 16-step DDIM. All three methods use this selection rule, and robot evaluation episodes are not used for checkpoint selection.

\subsection{Sketch Intervention Protocol}
\label{app:intervention}
The offline intervention in Figure~\ref{fig:intervention} fixes one handoff observation and the initial action noise, then replaces the sketch denoising trajectory with a donor trajectory at matching noise levels. We use EMA weights and deterministic 16-step DDIM. Donor sketches are expressed relative to the current joint configuration. They are replayed at each action-updating model call, so noisy sketch inputs retain the appropriate corruption level.

At inference step 120 of one episode, we generate 20 native front-route sketches. We take 20 additional front-route donors from the preceding observation and 20 back-route donors from a later observation in the same episode. Each set is paired with the same ten action noises. The different-route comparison uses the two transplanted sets; the same-route comparison uses native and transplanted front-route sets. The noise control varies action noise while holding the sketch fixed.

We measure the root mean squared Euclidean displacement between paired predicted tool positions at each action step, using forward kinematics with the fitted tool reference. This tests action dependence at a fixed observation, not closed-loop task success or collision clearance.

\textbf{Visualization.} Robot predictions are projected through UR3 forward kinematics into a calibrated recording camera separate from the policy cameras. Figure~\ref{fig:real_pred} uses a nominal flange offset of 0.20\,m; these overlays are qualitative. Projected coefficient markers are not image-space spline control points. In Figure~\ref{fig:intervention}a, paths are translated in the image plane to start at the visible gripper center for display only; quantitative panels retain the original tool-reference coordinates.